\documentclass[letterpaper, 10 pt, conference]{ieeeconf}  

\IEEEoverridecommandlockouts                              

\usepackage{times}
\usepackage{booktabs}
\usepackage{multicol}
\usepackage[bookmarks=true]{hyperref}
\usepackage{amsmath,amssymb,graphicx,MnSymbol}
\usepackage{subcaption}

\DeclareGraphicsExtensions{.png,.pdf,.jpg,.eps,.ps,.xcf}
\usepackage{algorithm}
\usepackage{algorithmic}

\newcounter{ALC@tempcntr}
\newcommand{\kvw}{\vec{w}^{k}}

\newcommand{\vw}{\vec{w}}

\newcommand{\s}{x}
\newcommand{\as}{x_{a}}
\newcommand{\ds}{x_{d}}

\newcommand{\tas}{x^{t}_{a}}
\newcommand{\tds}{x^{t}_{d}}
\newcommand{\tqdda}{\ddot{q}_a^t}
\newcommand{\tqddd}{\ddot{q}_d^t}

\newcommand{\tqd}{\dot{q}^t}

\newcommand{\tq}{q^t}

\newcommand{\qdda}{\ddot{q}_a}
\newcommand{\qddd}{\ddot{q}_d}
\newcommand{\qdd}{\ddot{q}}
\newcommand{\qd}{\dot{q}}
\newcommand{\q}{q}

\newcommand{\lossindirect}{\mathcal{L}_{\text{indirect}}(\vw)}
\newcommand{\lossdirect}{\mathcal{L}_{\text{direct}}(\vw)}
\newcommand{\klossindirect}{\mathcal{L}_{\text{indirect}}(\kvw)}

\newcommand{\klossjoint}{\mathcal{L}_{\text{joint}}(\kvw)}

\newcommand{\qdtaurbd}{\hat{\tau}_\text{rbd}(\ds)}
\newcommand{\qataurbd}{\hat{\tau}_\text{rbd}(\tas)}

\newcommand{\qdpolicy}{\pi(\tq,\tqd)}

\newcommand{\kdataindirect}{\mathcal{D}_\text{indirect}^{k}}
\newcommand{\kdatadirect}{\mathcal{D}_\text{direct}^{k}}
\newcommand{\kdatajoint}{\mathcal{D}_\text{joint}^{k}}

\newcommand{\symtaurbd}{\hat{\tau}_\text{rbd}}

\newcommand{\ttau}{\tau^t}
\newcommand{\taurbd}{\tau_\text{rbd}}
\newcommand{\tauapplied}{\tau_\text{total}}
\newcommand{\taufb}{\tau_\text{fb}}

\newcommand{\fid}{f_\text{id}}
\newcommand{\fiderr}{f_\text{iderr}}

\title{A New Data Source for Inverse Dynamics Learning}
\author{Daniel Kappler$^{\ast,1,2}$, Franziska Meier$^{\ast,1,2,4}$, Nathan Ratliff$^{2}$ and Stefan Schaal$^{1,3}$
\thanks{$^{\ast}$both authors contributed equally to this work}%
\thanks{$^{1}$AMD, MPI for Intelligent Systems, T\"ubingen, Germany}%
\thanks{$^{2}$Lula Robotics Inc, Seattle, USA}%
\thanks{$^{3}$CLMC Lab, University of Southern California, Los Angeles, USA}%
\thanks{$^{4}$RSE Lab, University of Washington, Seattle, USA}%
}

\begin{document}

\maketitle

\thispagestyle{empty}
\pagestyle{empty}

\begin{abstract}
Modern robotics is gravitating toward increasingly collaborative human
robot interaction.
Tools such as acceleration policies can naturally support the
realization of reactive, adaptive, and compliant robots. These tools
require us to model the system dynamics accurately -- a difficult
task. The fundamental problem remains that simulation and reality
diverge--we do not know how to accurately change a robot's state.
Thus, recent research on improving inverse dynamics models has been
focused on making use of machine learning techniques.
Traditional learning techniques train on the actual realized
accelerations, instead of the policy's desired accelerations, which is
an indirect data source.
Here we show how an additional training signal -- measured at the
desired accelerations -- can be derived from a feedback control
signal. This effectively creates a second data source for learning
inverse dynamics models.
Furthermore, we show how both the traditional and this new data
source, can be used to train task-specific models of the inverse
dynamics, when used independently or combined.
We analyze the use of both data sources in simulation and demonstrate its
effectiveness on a real-world robotic platform. We show that our
system incrementally improves the learned inverse dynamics model, and
when using both data sources combined converges more consistently and
faster.
\end{abstract}

\section{Introduction}
%
Achieving reactive and compliant behavior is a cornerstone of robotic
applications involving safe interaction with humans. A promising
avenue for realizing reactive behaviors is the representation of
motions through acceleration policies. Unfortunately, tracking
accelerations is hard when we do not have an accurate model of the
inverse dynamics of the system. Without a precise model of the
systems dynamics, we typically resort to tracking desired
trajectories--potentially generated by integrating the acceleration
policy--employing feedback control to reject modeling errors of the
dynamics. After careful tuning of the controller gains, this usually
results in good tracking performance. However, we have to trade off
compliancy and reactiveness of our controller against accuracy.
Thus, considerable effort has been put into developing machine
learning methods that can learn or improve inverse dynamics models
\cite{Vijayakumar2000, Nguyen-Tuong2008, Gijsberts2013, Meier2014a}.
These approaches attempt to identify a global inverse dynamics model,
but collecting data that covers the full state-space is typically not
considered a viable approach for high-dimensional systems.
Furthermore, when considering motions with object interaction, such as
pick and place tasks, learning one global model becomes even more
involved, if not impossible, since the model has to be a function of
contact and payload signals.
Thus online learning has been a focus in these settings.
However, online learning in this setting remains a challenge.
The key difficulties are computationally efficient learning of models
that are flexible enough to capture the non-stationary data of inverse
dynamics mapping, and doing so on streaming data that is highly
correlated.

To circumvent the issues of the aforementioned methods we follow the
path of learning {\em task-specific\/} (error) models
\cite{jamone2014incremental, toussaint2005learning,
  petkos2006learning, wolpert1998multiple}.
This allows us to iterate collecting data specific to a task, learn
an error model in offline fashion and then apply the learned model
during the next task execution.
While this approach is iterative by nature, we aim at making it as
data-efficient as possible, such that only few iterations are
required, while achieving consistent convergence in the error model
learning process.
Task specific models do not mitigate the problem of using payload
signals or contextual information when trying to build a global model.
However, it simplifies the overall global problem into two
subproblems, finding a task specific inverse dynamics model and
detecting which task model to use.
In this context a typical issue of traditional inverse dynamics
learning approaches is that they actually learn the (error) model
slightly off the desired trajectory. This comes from the fact that the
data points used to learn the model are based on the actual achieved
accelerations instead of the desired commanded accelerations. Thus if
tracking is bad the collected data points are slightly off the
desired trajectory for which we wanted to identify the inverse
dynamics model. For instance, one extreme case of bad tracking is
encountered when the system cannot overcome static friction. In that
case, no useful data for inverse dynamics model learning is generated.
Traditionally, this is circumvented by increasing the gains of the
feedback control term, at the cost of compliancy. Alternatively, this
issue is addressed by our recent work \cite{ratliff2016doomed}, which
employs a \emph{direct} loss function, minimizing the error between
desired and actual accelerations, to learn feedback terms online.

In this work, we explore the intuition that feedback control can be
viewed as an online technique to compensate for errors in a given a
priori inverse dynamics model, as discussed in
\cite{ratliff2016doomed}. The feedback terms, compensate for errors
between what a given model predicts and what we actually need.
Therefore they naturally act as a convenient source of training data.
We show how the \emph{direct loss} on accelerations, as presented in
\cite{ratliff2016doomed}, can be transformed into a loss on inverse
dynamics torques, measured at desired accelerations. As a result, we
now have two training data sources: the traditional inverse dynamics
training data points measured at actual accelerations, and this new
training signal measured at commanded accelerations. We show how this
additional data source leads to more consistent convergence of the
task-specific inverse dynamics learning process.
In the following, we first review the inverse dynamics learning
problem in Section~\ref{sec:bg} and discuss the differences between
indirect and direct learning in this setting. In
Section~\ref{sec:offline_and_online} we show how to combine the direct
loss with the traditional indirect loss for inverse dynamics learning,
followed by an overview of our complete task-specific learning
approach in Section~\ref{sec:task_specific_learning}. Finally, we
evaluate our proposed system in Section~\ref{sec:Experiments}.
%

\section{Background}
\label{sec:bg}
%
Tracking desired accelerations with low feedback controller gains
requires an accurate inverse dynamics model. The dynamics of a
classical dynamical system can be expressed as
\begin{align}
	\tau = M(\q)\qdd + h(\q, \qd)
\end{align}
where $\q, \qd, \qdd$ denote the joint positions, velocities and
accelerations, $M$ is the inertia matrix and $h$ collects all the
modeled forces such as gravitational, Coriolis, centrifugal forces,
viscous and Coulomb friction. When possible, inverse dynamics
approaches \cite{Craig1987} model the system dynamics via the rigid
body dynamics (RBD) equation of motions. Then, given sufficiently rich
data, the RBD parameters can be identified using linear regression
techniques \cite{An1985}, resulting in the approximate RBD dynamics
model
\begin{align}
	\hat{\tau}_\text{rbd} = \hat{M}(\q) \qddd + \hat{h}(\q, \qd).
\end{align}
with approximate $\hat{M}$ and $\hat{h}$. This has been extended in
\cite{Atkeson1985}, to additionally estimate payloads.

Unfortunately, the RBD model typically is not flexible enough to
capture all non-linearities of the actual systems dynamics. As a
result, the estimated RBD model is generally only a rough
approximation. Thus, when attempting to track desired accelerations
$\qddd$ with $\hat{\tau}_\text{rbd}$ we achieve actual accelerations
$\qdda$ differing from $\qddd$. Specifically when
$\hat{\tau}_\text{rbd}$ is applied on the real system, with the true
unknown dynamics model $M, h$, we can express the actual accelerations
$\qdda$ as
\begin{align} \label{eqn:ActualRBDAccelerations}
	\qdda &= M(\q)^{-1}[ \hat{\tau}_\text{rbd} - h(\q, \qd)]\\
	&= M(\q)^{-1}[ \big(\hat{M}(\q) \qddd + \hat{h}(\q,
\qd)\big) - h(\q,\qd)].
\end{align}
Note, if our estimated model $\hat{M}, \hat{h}$ were accurate, this
expression would evaluate to $\qdda = \qddd$. However, this is
typically not the case on real systems and because of this a feedback
term $\taufb$ is required. This feedback term measures the error made
and adds a corrective term to ensure accurate tracking. Traditionally,
this feedback term is realized through PID control. The higher the
gains, the better the tacking, at the cost of compliancy.
\subsection{Learning Inverse Dynamics Models}
To this end, various approaches to learning either the full inverse
dynamics or an error model have been proposed. When learning an error
model the total torque command is then a combination of any existing
approximate model ($\symtaurbd$), an error (torque) model
($\fiderr$) and a feedback term ($\taufb$),
\begin{align} \label{eqn:CorrectedModel}
\tauapplied = \symtaurbd + \fiderr + \taufb.
\end{align}

One of the key challenges of inverse dynamics learning is
computational efficiency. Predicting with learned models needs to be
feasible within the real-time constraints of the systems consuming
torque commands. Furthermore, it is typically assumed that the inverse
dynamics mapping is non-stationary and can change over time. Thus,
approaches that can incrementally learn and are computationally
efficient enough for real-time deployment
\cite{Vijayakumar2000,Nguyen-Tuong2008,
  Gijsberts2013,Meier2014a,jamone2014incremental,Meier2016drifting}
form one of the main research directions within the topic of inverse
dynamics learning. However, learning globally valid models robustly on
highly correlated data streams remains a challenge. More in depth
discussion about the challenges and existing approaches can be found
in \cite{sigaud2011online,Nguyen-Tuong2011}.

Some robustness can be achieved by using analytical (parametric)
models such as RBD models as priors, and learning an error model on top
of that \cite{Nguyen-Tuong2010, Cruz2012online,
  camoriano2016incremental}. Such approaches can revert to this prior
knowledge, when the algorithm determines that the error model fit is
uncertain.

Because of the difficulties of learning a globally valid model, some
research has moved towards learning task/context specific inverse
dynamics models \cite{wolpert1998multiple, toussaint2005learning,
  petkos2006learning, petrivc2014online, jamone2014incremental,
  calandra2015learning, christiano2016transfer}. In this setting it is
feasible to collect task relevant data for offline learning, therefore
simplifying the learning process. 
However, this comes at the cost of
having to detect the correct context at run time such that the correct
task model can be chosen. Our work, fits into this category, with the
focus on effectively learning one task model.

Finally, learning inverse dynamics models is typically motivated by
being able to use low-gain feedback control. However, traditional
inverse dynamics learning approaches initially need high enough gains
to achieve good tracking, such that relevant data is being generated.
Little work has been done towards automatically lowering the feedback
gains once a good model has been learned. An exception is work
presented in \cite{alberto2014computed} which allows for variable
gains, using high gains, when the model is uncertain about its
predictions and low gains when it is certain.
All of these methods learn inverse dynamics models on only the
indirect data source measured at actual accelerations. In this work, we
make use of two different data sources stemming from indirect and
direct learning approaches to train inverse dynamics models. 

Thus, before going into the details of our proposed approach, we
discuss the notion of \emph{indirect} and \emph{direct} learning and
present related work within that context.
\subsection{Indirect vs Direct Learning of Inverse Dynamics}
\label{sec:ilid}
As mentioned above, most of the recent work on inverse dynamics learning can
be classified as \emph{indirect} learning methods. The objective
function that these methods optimize is given as
\begin{align}
	\lossindirect = \sum_{(\as, \tauapplied) \in \mathcal{D}} || \tauapplied - f(\as; \vw)|| \label{eq:loss_indirect}
\end{align}
where the input data point
$\as = (\q,\qd,\qdda)$ is a combination of the state $(\q, \qd)$ and
the actual accelerations $\qdda$. The output value $\tauapplied$ is
the corresponding applied torque that achieved the accelerations
$\qdda$. Here, the function that encodes the mapping from $\s$ to
torques is defined as $f(\s; \vw)$, where $\vw$ are the open
parameters of the chosen model.

Online, the robot attempts $\qddd$ from state $(\q,\qd)$. It
calculates torque $\tauapplied$ and observes true accelerations
$\qdda$.
Rather than training at input point $\ds = (\q, \qd, \qddd)$, the data
point $\as = (\q, \q, \qdda)$ is used with target $\tauapplied$.
The drawback to indirect learning is that convergence might be slow
since the data is off the trajectory/distribution we want to optimize
for.
This is especially difficult when $\ddot{q}_a = 0$ due to static
friction since many torques map to this value (i.e. the inverse
function is not one-to-one at this point).

Alternatively, it was shown in \cite{ratliff2016doomed} that it is
possible to directly measure the gradient of the acceleration error of
a system, which enables a new class of direct online learning
algorithms.
This approach aims to directly minimize the error between desired and
actual accelerations, by optimizing the following {\em direct\/} loss
function, for every pair of desired and actual accelerations,
\begin{align}
	\lossdirect &=  ||\qddd - \qdda(\vw)||_M^2 \label{eq:loss_direct}\\\nonumber
	&= ||\qddd - M^{-1}[ \hat{\tau}_\text{rbd} + \fiderr - h]||_M^2 %
\end{align}
where the actual accelerations $\qdda$ are a function of the error
model $\fiderr$.
Note, traditional direct adaptive control methods
\cite{AdaptiveControl2008, NAKANISHI20041453} have similar motivations -- they use
Lyapunov techniques to derive controllers that adjust the dynamics
offset model with respect to some reference signal. 
In
\cite{ratliff2016doomed} we show how to effectively perform online
learning on this objective leveraging well established online gradient
descent techniques.
However, this simple feedback term does not capture any structure of
the error model and requires a relatively high learning rate to
account for payload changes. In \cite{Meier2016drifting} we thus use
\cite{ratliff2016doomed} as a feedback term on acceleration errors and
use the indirect loss function to learn a drifting Gaussian process to
model larger structured inverse dynamics modeling errors. Note,
\cite{Meier2016drifting} combines the direct and indirect learning as
two separate learning processes with two different purposes: direct
learning of an online adaptive feedback term and indirect learning of a
locally valid inverse dynamics error model to capture larger errors.
Also, \cite{Meier2016drifting} is a task independent online learning
approach and theoretically applicable anywhere throughout the state
space. However, on the flip side it retains no memory of previously
learned error models, and thus is not able to improve over time.

Our work presented here is orthogonal to our previous work: We show
that we can use direct and indirect learning within the same learning
process of task-specific (feedforward) inverse dynamics (error)
models, which can improve over time.
\section{Combining Indirect and Direct Learning}
\label{sec:offline_and_online}
%
To be able to learn the state-space dependent structure of the inverse
dynamics modeling errors, we assume that the error model is identifiable
in the space of $\s = (\q,\qd,\qdd)$.
Furthermore, since every task execution may slightly vary, we also
follow an incremental learning process, meaning that each task
execution generates a new training data set that can be used to update
and improve our error model.
Thus, our error models are indexed by $k$, indicating the $k^{th}$ learning
iteration. For $k=0$, meaning that no error model exists for the task
at hand, we simply assume $\fiderr^0(\s; \vw^0) = 0$. 
Given this, the total torque applied to the system is the approximate
rigid body dynamics model $\qdtaurbd$ (if available) plus an offline
learned error model $\fiderr^{k}$ and a feedback term $\taufb$:
\begin{align}
  \tauapplied &= \qdtaurbd + \fiderr^{k}(\ds; \vw^{k}) + \taufb.  \label{eq:tauapplied}
\end{align}
Here we show how the direct and indirect loss functions can be
combined into one loss function that uses two different data sources.
In order to do so we 1) discuss the loss functions in the context of
offline error model learning, 2) show that the two loss functions
create two different training signals for the error model, and 3) use
this result to combine direct and indirect learning.
%
\subsection{Indirect Loss Function}
\label{sec:iidl}
We start out with discussing the details of learning an error model
with an indirect loss.
We compute the torque command based on the current state $(\q, \qd)$
and desired accelerations $\qddd$, apply this torque, and then measure
actual accelerations $\qdda$. Now we know what torque command achieves
these measured accelerations and can use this data point to learn an
inverse dynamics model. We collect all of these data points over the
course of one task execution, for $t=1 \dots T$, such that we have $T$
data points to learn parameters $\vw^{k}$, initialized with the
parameters $\vw^{k-1}$.

In the indirect formulation, we try to optimize the parameters
$\vw^{k}$ such that the difference between the applied torque
$\tauapplied$ and the inverse dynamics model $\fid$ at $\as$ is
minimized:
\begin{align}
  \klossindirect = \sum\nolimits_{t=1}^{T} \|\tauapplied^t - \fid(\tas; \vw)\|^2
\label{eq:iidl}
\end{align}
Here we would like to utilize an approximate rigid body dynamics model
(if available) and learn an error model $\fiderr$ in order to optimize the
$\fid$ model.
Notice, our approach does not require a rigid body dynamics model,
all derivations hold when assuming a constant model $\hat{\tau}_\text{rbd} := 0$ as well.
In this case we would learn the full inverse dynamics model, not using any
domain specific knowledge.
To compute what the RBD error is at input $\tas$, we have to
evaluate $\symtaurbd$ at $\tas$ and subtract it from the total torque
applied $\tauapplied$, such that, in the $k^{th}$ learning iteration,
we optimize
\begin{align}
  \klossindirect = \sum\nolimits_{t=1}^{T} \|\tauapplied^t - \qataurbd - \fiderr^k(\tas; \vw^k)\|^2\nonumber
\end{align}
Thus, using the indirect learning approach we optimize
$\fiderr^{k}(\s; \vw^{k})$ on the following data set
\begin{align}
	\kdataindirect = \{ x^t \leftarrow \tas ,
  y^t  \leftarrow  \tauapplied^{t} - \qataurbd \}_{t=1}^T. \label{eq:data_indirect}
\end{align}
The quality of this training data set depends on how well
we have tracked the task policy or trajectory.
With accurate tracking behavior,
one learning run should already give us a good approximation of the
modeling errors. However, if tracking is bad, it very well may be that
we require several learning iterations to estimate a good error model.
%

%
\subsection{Direct Loss Function}
\label{sec:didl}
%
To overcome the limitations of the indirect learning process,
\cite{ratliff2016doomed} proposes to use a direct loss
(Eq.~\ref{eq:loss_direct}) to learn modeling errors.
Here we use this loss in acceleration space
\cite{ratliff2016doomed}, to derive an additional data source for
inverse dynamics learning.

We start out with Eq.~\ref{eq:loss_direct}, drop the weighting of the
acceleration error by the inertia matrix $M$, and instead multiply the
accelerations with $M$
\begin{align}
	\lossdirect &= \sum\nolimits_{t=1}^T \|M\tqddd - M\tqdda(\vw^{k}) \|^2 \\\nonumber
              &= \sum\nolimits_{t=1}^T \|(M\tqddd + h) - M\tqdda(\vw^{k}) -h \|^2\nonumber 
\end{align}
where we have also added and subtracted $h$.
The true dynamics model $M,h$ is never evaluated in our loss
formulation, it is merely used to derive the direct loss formulation
as shown in the following.
We can now summarize the first term as the true rigid body dynamics
model $\taurbd(\tqddd)$, evaluated at the desired accelerations, and we
expand $\tqdda(\vw^{k})$ as follows
\begin{align}
	\lossdirect &= \sum\nolimits_{t=1}^T \|\taurbd(\tqddd) - M\tqdda(\vw^{k}) -h\|^2\\\nonumber 
	 &= \sum\nolimits_{t=1}^T\|\taurbd(\tqddd) - M M^{-1}[f_{id}(\tqddd; \vw^{k}) -h] -h\|^2\\\nonumber 
	 &= \sum\nolimits_{t=1}^T\|\taurbd(\tqddd) - (\symtaurbd(\tqddd) + \fiderr^k(\tds; \vw^k))\|^2\\\nonumber
	 &= \sum\nolimits_{t=1}^T\|(\taurbd(\tqddd) - \symtaurbd(\tqddd)) - \fiderr^k(\tds; \vw^k)\|^2 
\end{align}
where $f_{id}(\tqddd)$ represents the state based rigid body dynamics and
error model without a feedback term which should ideally be zero.
We now have transformed the loss on accelerations to a loss on torque
commands at the input point $\tds$. Note that this transformed loss
intuitively means that we want to minimize the difference between our
error model $\fiderr^k(\tds; \vw^k)$ and the true modeling error
$\tau_\text{error}^{t} = (\taurbd(\tqddd) - \symtaurbd(\tqddd))$ at input
$\tds = (\tq, \tqd, \tqddd)$. While intuitively pleasing, we unfortunately do
not have access to the true modeling error $\tau_\text{error}^{t}$.
However, we can get an estimate of the modeling error
$\hat{\tau}_\text{error}^t = \taufb^t + \fiderr^{k-1}(\tds;
\vw^{k-1})$ by combining the feedback term $\taufb^t$ and the error
model $\fiderr^{k-1}$ from the previous task execution (
similar to feedback error learning \cite{NAKANISHI20041453}), which results
in the following loss
\begin{align}
	\lossdirect &= \sum\nolimits_{t=1}^T \|\hat{\tau}_\text{error}^{t} - \fiderr^{k}(\tds; \vw^{k})\|^2 
\end{align}

Similar to the indirect learning we can now construct a data set
\begin{align}
  \kdatadirect = \{ 
    x^t \leftarrow \tds,
  y^t \leftarrow \taufb^t + \fiderr^{k-1}(\tds; \vw^{k-1}) \}_{t=1}^T \label{eq:data_direct}
\end{align}
which can be used to learn or update the new error model
$\fiderr^{k}$. 
We now receive data points directly on the
desired accelerations. 
However, also this data set's quality depends on
tracking accuracy. With low feedback gains, the initial $\taufb$ may
not really capture the errors very well, such that the first learning
iteration may only capture part of the modeling errors.

Thus, with low feedback gains, we may require multiple learning
iterations to learn an accurate error model.
Whereas increasing the feedback gains $g$ would lead to improved
tracking, increasing the fidelity of the data, at the cost of
compliancy.
Here we simply propose to do both: use $g_\text{low}$ to compute the
feedback terms $\taufb(g_\text{low})$ which are sent to the system,
and use $g_\text{high}$ to compute feedback terms
$\taufb(g_\text{high})$ which are sent to the learner.
Notice, the feedback term using $\taufb(g_\text{high})$ is never
applied on the system, thus, we maintain a very compliant system while
obtaining better error data for the portion of the state space reached
with the $g_\text{low}$. This can be helpful to break stiction or
counteract high friction with $\fiderr$ after fewer iterations which
otherwise would not be possible with traditional inverse dynamics
learning approaches and low gains.

\subsection{Joint Inverse Dynamics Learning}
\label{sec:jidl}
The key insight of this paper is that we can use the feedback term
as an error estimate for the desired accelerations.
Thereby the error model learning problem has two data sources
{\em indirect\/} and {\em direct\/}.
Both exhibit the same structure to optimize the error model.
Hence, we can formulate a joint function approximation problem of the form:
\begin{align}
\klossjoint = \sum_{(\s,y) \in \kdatajoint}
  \| y - \fiderr^{k}(\s ; \kvw) \|
  \label{eq:jidl}
\end{align}
where data points for the actual accelerations $\tqdda$ for every
timestep $t$ can be used as well as data points for the desired
accelerations $\tqddd$ as described by
\begin{align}
  \kdatajoint = \kdatadirect \cup \kdataindirect. \label{eq:data_joint}
\end{align}
%

\section{Task Specific Inverse Dynamics Learning}
\label{sec:task_specific_learning}
%
Implementing our approach required several design decisions on the
levels of motion generation, control and learning. Here we will give a
short overview of our design choices, and give an algorithmic overview
of our iterative approach to learning task-specific inverse dynamics
models.

On the motion generation level, we assume that a kinematic policy for
our task is provided, meaning that we can obtain desired accelerations
$\tqddd$ for every state $\tq,\tqd$ relevant to our task. In
particular, we use kinematic Linear Quadratic Regulators (LQRs)
\cite{OptimalControlEstimationStengel94} to provide us with the
acceleration policy that is being executed. On the control level we
use two types of feedback controllers: traditional PID control and the
recently introduced adaptive feedback learning (DOOMED
\cite{ratliff2016doomed}).
When tuned sufficiently well, both approaches can result in good
tracking performance. We also have to choose a function approximator
for the error model.
Because we are learning the error model offline, learning speed is not
our main concern. 
However, prediction speed is important since the models need to be
evaluated at 1000Hz (hard real-time) on our robotic platform.
We choose a simple feedforward neural network which is capable of
learning nonlinear mappings and predictions require only a simple
feedforward pass.

Finally our iterative learning approach (Algorithm~\ref{alg:main}) can
be summarized as follows:
\begin{enumerate}
\item task execution: Run task with approximate RBD model,
  feedback control, and feedback error model if existent (line 5).
  Construct data set's with two different sources of information
  (equation~\eqref{eq:data_indirect} (line 10)
  and~\eqref{eq:data_direct}) (line 11) during the task execution, as
  shown in Algorithm~\ref{alg:main}.
\item learning phase: Update the error model function approximators
  based on new data points, or construct and initialize the error
  models if none exist (line 14).
\item repeat the task.
\end{enumerate}

\setlength{\textfloatsep}{4pt}
\begin{algorithm}[t]
\caption{Execute Task and Collect Data}\label{alg:main}
\begin{algorithmic}[1]
\REQUIRE $\fiderr^{k-1}$, $\qdpolicy$, system($\ttau$), $\q^{0},\qd^{0}$
\STATE $\kdatajoint = \emptyset; t=0$
\WHILE{ not converged }
\STATE $\qddd^t = \qdpolicy$
\STATE $\s^t = (\q^t, \qd^t, \qddd^t)$ 
\STATE $\ttau = \qdtaurbd + \taufb^t + \fiderr^{k-1}(\tds; \vw^{k-1})$
\STATE $t = t+1$
\STATE $\tq = $ system($\tau^{t-1}$)
\STATE $\tqd,\qdda^{t-1} = $ finiteDiff($\tq$,$\q^{t-1}$)
\STATE update feedback term $\taufb^t$
\STATE $\kdatajoint = \kdatajoint \cup (\as^{t-1},
\tau^{t-1} - \hat{\tau}_{\text{rbd}}(\as^{t-1}))$
\STATE $\kdatajoint = \kdatajoint \cup (\ds^{t-1}, \taufb^{t-1} +
\fiderr^{k-1}(\ds^{t-1}; \vw^{k-1}))$
\STATE 
\ENDWHILE
\RETURN optimize($\fiderr^{k}$, $\kdatajoint$)  
\end{algorithmic}
\vspace{-0.1cm}
\end{algorithm}
\setlength{\textfloatsep}{6pt}
As we empirically show in the Section~\ref{sec:Experiments},
exploiting both sources of information allows us to obtain a better
fit of the error model with less task iterations, compared to using
a single data source.
%
%
\section{Experiments} \label{sec:Experiments}
%
We evaluate our approach in two different settings. First, we
analyze the proposed usage of indirect and direct data sources in
order to learn a task specific inverse dynamics error model on a 2D
simulation.
This allows us to extensively test characteristics of the learning
problems based on the different data sources, using the same function
approximator, under various simulated noise levels, stictions, frictions,
and a wrong RBD model.
This evaluation is focused on investigating the importance and influence of the different data sources rather then the function approximator itself.
Second, we report results on task specific inverse dynamics learning
using both data sources on the KUKA lightweight arm of our robotic platform shown in Figure~\ref{fig:apollo_result_mse}.
We start out describing the evaluation of our 2D simulation setting.
%
%

\subsection{Simulation Experiments}
\begin{figure*}
    \centering
    \begin{subfigure}{.24\textwidth}
      \centering
      \includegraphics[width=0.99\linewidth]{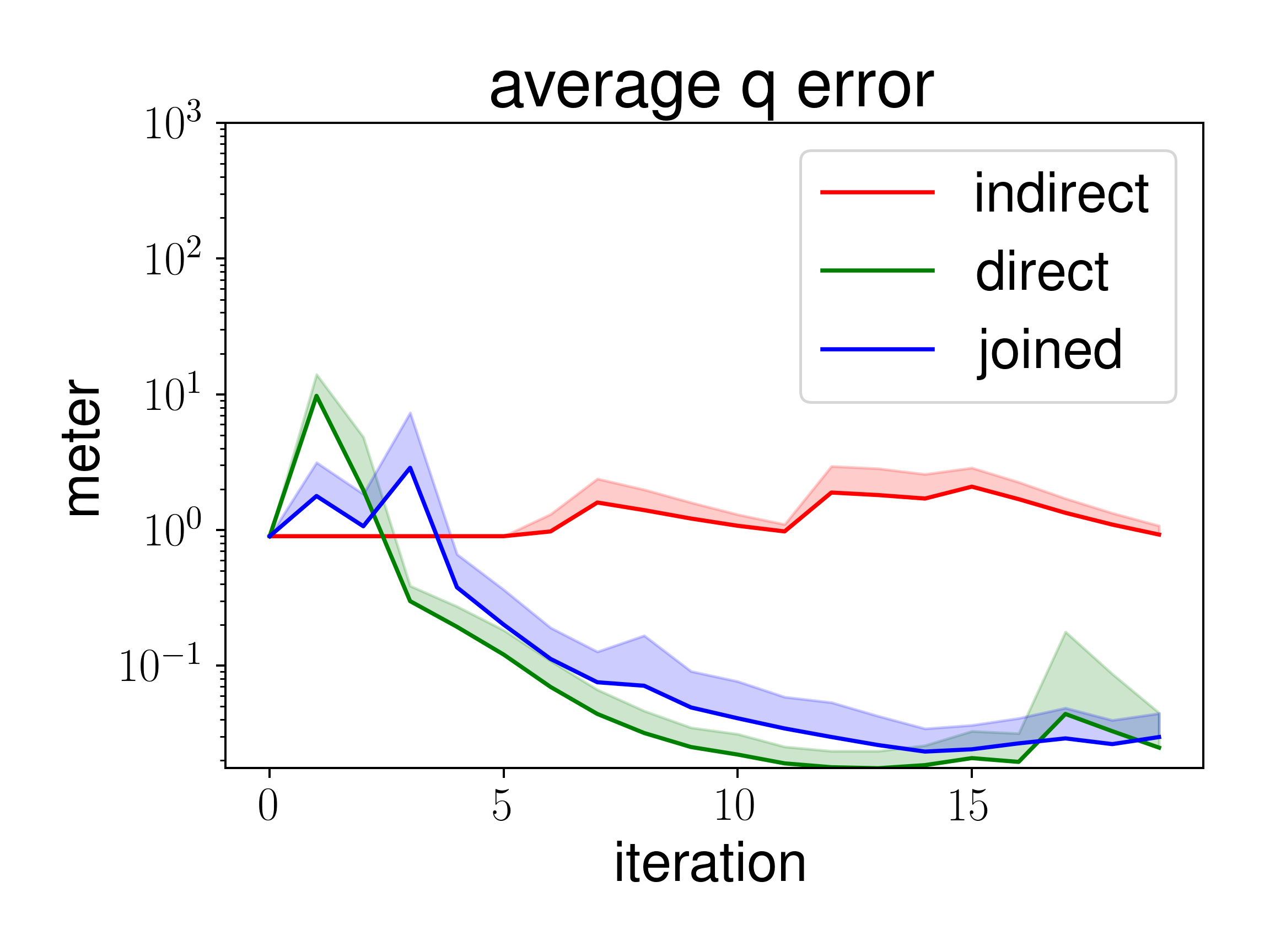}%
      \caption{PID, low system noise}
    \end{subfigure}
    \begin{subfigure}{.24\textwidth}
      \centering
      \includegraphics[width=0.99\linewidth]{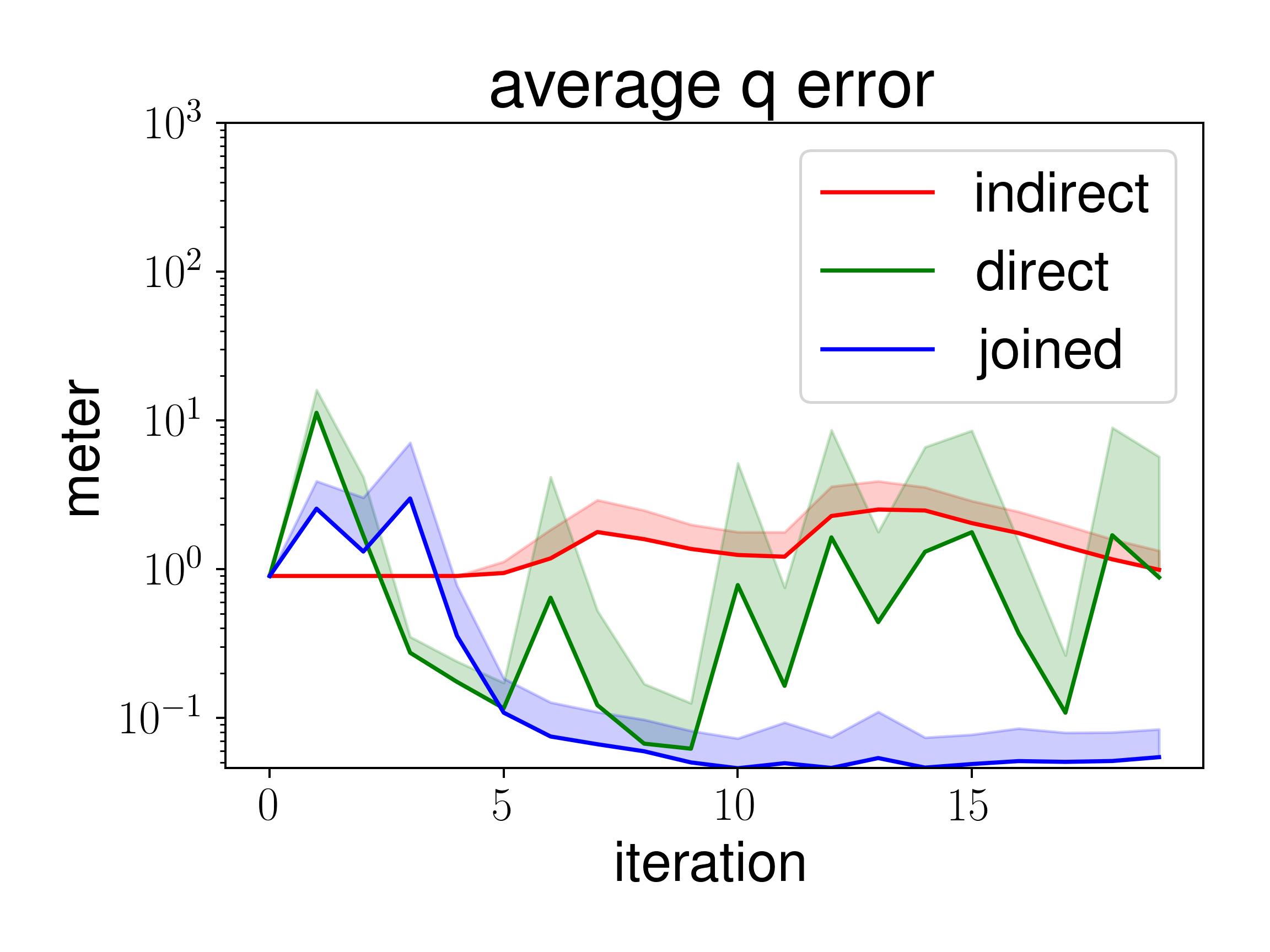}%
      \caption{PID, very high system noise}
    \end{subfigure}
    \begin{subfigure}{.24\textwidth}
      \centering
      \includegraphics[width=0.99\linewidth]{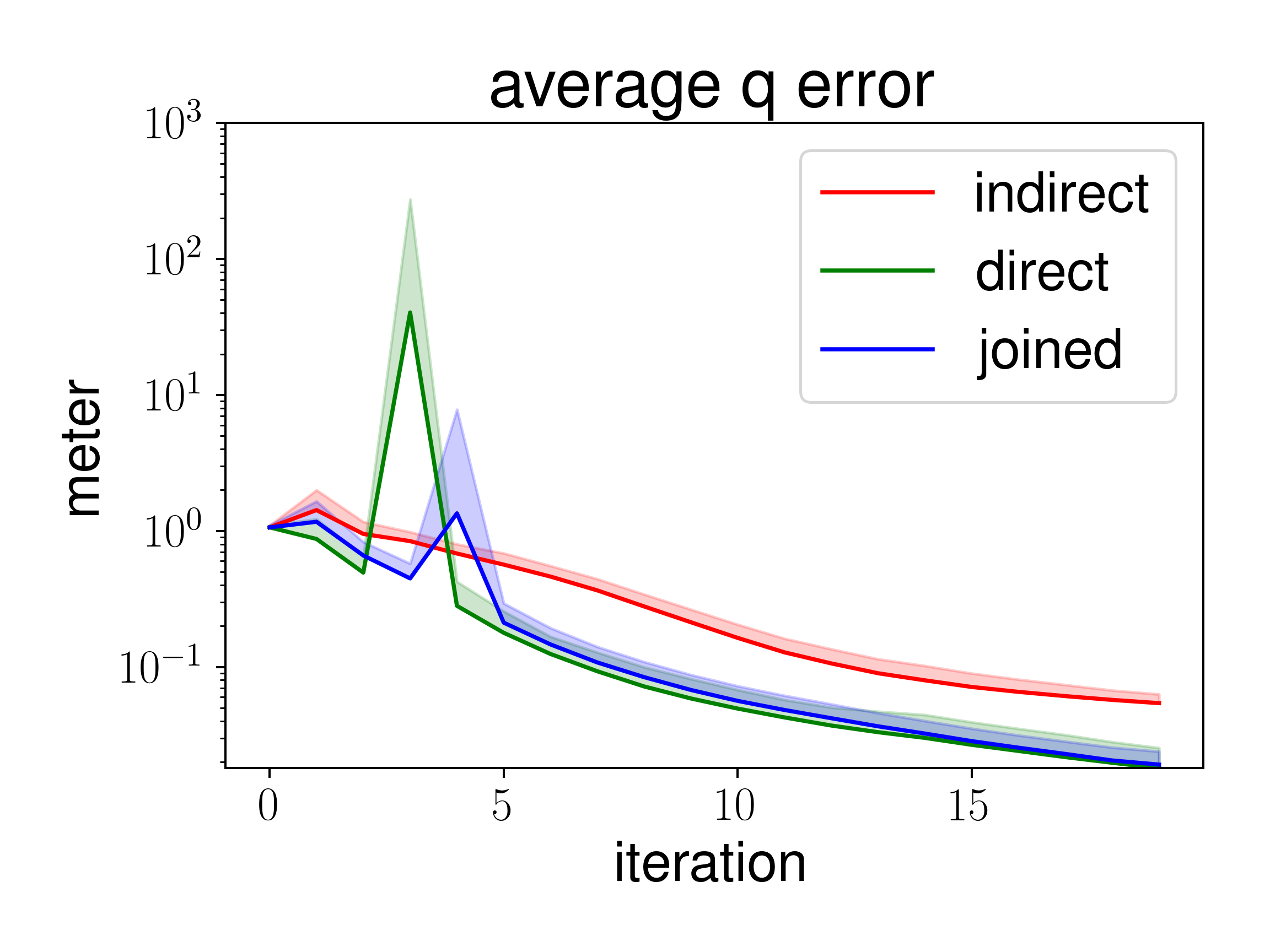}
      \caption{doomed, low system noise}
    \end{subfigure}
    \begin{subfigure}{.24\textwidth}
      \centering
      \includegraphics[width=0.99\linewidth]{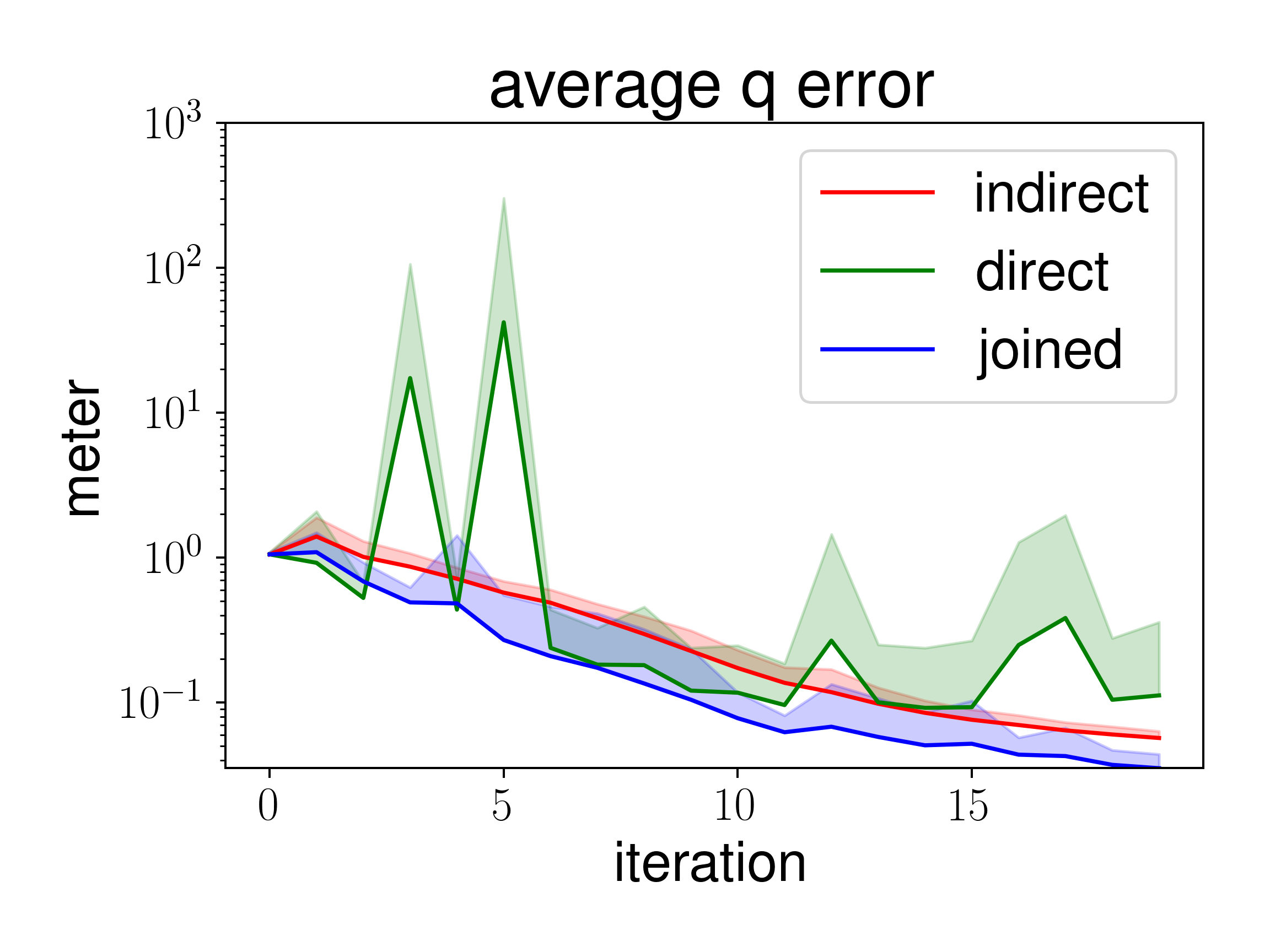}
      \caption{doomed, very high system noise}
    \end{subfigure}
    \caption{\small
      Hand-picked simulation runs to illustrate
      (dis)-advantages of the \emph{direct} or \emph{indirect} data
      sources. On the top row we show position tracking error as a
      function of learning iterations, for both low-gain PID and low-gain
      DOOMED feedback control on systems with low and very high noise. In
      low noise settings, for both PID and DOOMED, the direct data
      source leads to improved position tracking with increasing
      number of learning iterations. However, when using the indirect data
      source the NN cannot capture the error model, and thus position
      tracking does not improve at all (with PID) or more slowly (with
      doomed) over time. In the very high noise setting, the learning
      convergence with indirect data source is comparable to the low
      noise setting, but when using the direct data source alone, we
      see erratic tracking convergence. The direct data is affected
      more by the noise. For both PID and DOOMED, the position
      tracking performance converges more consistently when using the
      joint data set.
	\vspace{-0.4cm}}
	\label{fig:illustration_direct_vs_indirect}
\end{figure*}
%
%
\subsubsection{System description}
This evaluation is based on a simple 2D example \cite{idsim} of a
system for which the approximate RBD dynamics model differs
drastically from the true dynamics.
In this simulated system the true mass is set to $\mathbf{M} = 5 I$,
while the approximate mass is assumed to be $\mathbf{M} = 0.5 I$. The
system attempts to realize a simple acceleration policy, defined as a
PD-controller with desired state at $\q_\text{des} = (1, 1)$ and 
initial state $\q_\text{init} = (0, 0)$ and
$\qd_\text{init} = (0, 0)$.
Furthermore, we simulate the true system to experience friction and
stiction which is not modeled by the approximate RBD 
model.
Finally we also add sensing noise to the position trajectories to
mimic noisy sensor measurements.
The source code of our example simulation with all parameters,
friction and stiction models used for the experiments can be found at
\cite{idsim}.
\subsubsection{Details of error model learning}
Every experiment involves running the iterative learning process for
$20$ iterations. After each cycle a neural network is trained on the
collected data and is then used to predict the modeling error in the
next cycle.
The error model $\fiderr$ is learned via a neural network structure which
consists of fully connected layers (200, 100, 50, 20, 1) with
non-linearities (prelu \cite{he2015delving}) after every layer except
the last. 
We optimize one network per simulated joint of our system.
The neural network is trained on the indirect data set
(equation~\eqref{eq:data_indirect}), the direct data set
(equation~\eqref{eq:data_indirect}), and as proposed in this paper the
joint data set (equation~\eqref{eq:data_joint}).
\subsubsection{Experimental Setup}
To evaluate the use of the data source variants (direct, indirect or
joint) in a principled manner, we simulate various system conditions:
\begin{itemize}
\item 4 maximum sensing noise levels in meters are reported: low
  (0.0001), medium (0.0005), high (0.007), very high (0.008)
\item 2 friction levels: medium and high  
\item 2 stiction levels: medium and high 
\end{itemize}

across different hyper-parameter settings:
\begin{itemize}
\item 2 different number of training epochs of the neural network training per
  task iteration are evaluated (20, 50).
\item 2 different gain settings (applied) for DOOMED and PID: low and
  high. The PID gains have been tuned such that even without error
  model we achieve convergence to the goal. Low gains are one order of
  magnitude lower.
\end{itemize}

The friction model is discontinuous and changes throughout
the state space.
The feedback term for learning is exponentially
filtered with the same value (0.1) for all experiments.
This is possible since the learning feedback term
is not applied to the system.

For the detailed noise, friction and stiction models as well as the
exact parameters to replicate the experiments please check out \cite{idsim}.
For each data source variant (direct, indirect, joint) a total of 16
system combinations were executed, $10$ times each, with
different random seeds.
A total of 1280 experiments were performed to cover the different
hyper-parameter settings.
To make results comparable, the random seed was kept consistent across
the data source variants.
For each of these runs we record the average magnitude of the applied
feedback torque, the desired and actual acceleration, as well as the
desired and actual position.

\subsubsection{Illustration of Indirect vs Direct}
We start out with illustrating some hand-picked scenarios that
showcase the differences of learning on indirect vs direct data in
Figure~\ref{fig:illustration_direct_vs_indirect}. We choose examples
with the same parameter settings, and illustrate the difference when
going from low to high noise. For the chosen examples, the learning
rates/feedback gains were set to the low value, such that friction and
stiction was not so easily overcome with feedback terms alone. This
has the effect that for low-gain PID control we basically see no
improvement in position tracking over time when using the indirect
data only. With DOOMED, some improvement can be observed (with
indirect data) - but at a slower rate compared to using direct data.
The noise level does not affect the indirect learning process much.
However, it affects the learning on direct data. In the low noise setting
we observe how position tracking improves over time, but in the high
noise setting this is not true. However, when combining both data
sources, we get consistently good convergence of position tracking
performance, even with low-gain feedback control.

\subsubsection{Extensive Evaluation}
\begin{figure*}
    \centering
    \begin{subfigure}{.24\textwidth}
      \centering
      \includegraphics[width=0.99\linewidth]{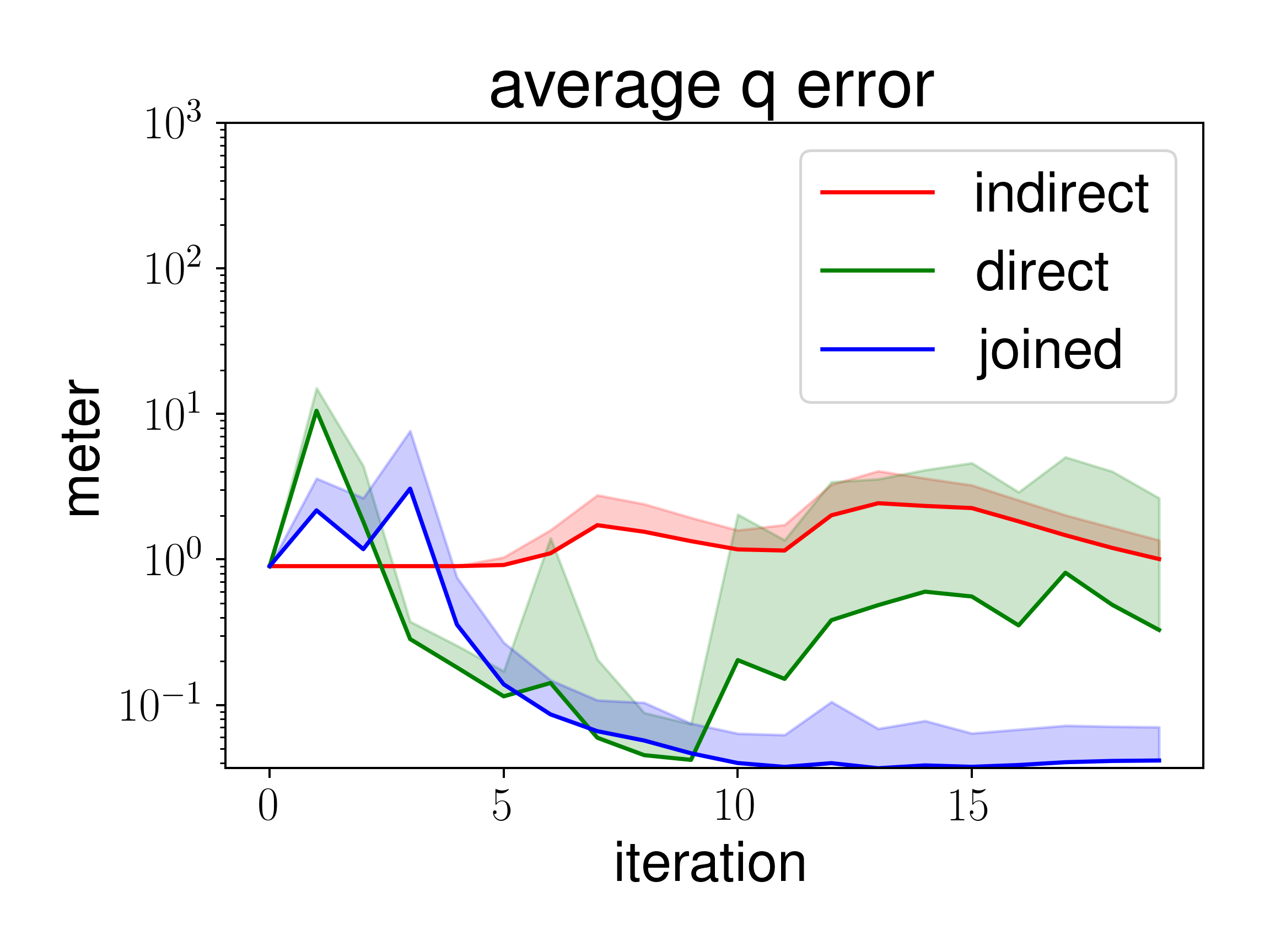}%
    \end{subfigure}
    \begin{subfigure}{.24\textwidth}
      \centering
      \includegraphics[width=0.99\linewidth]{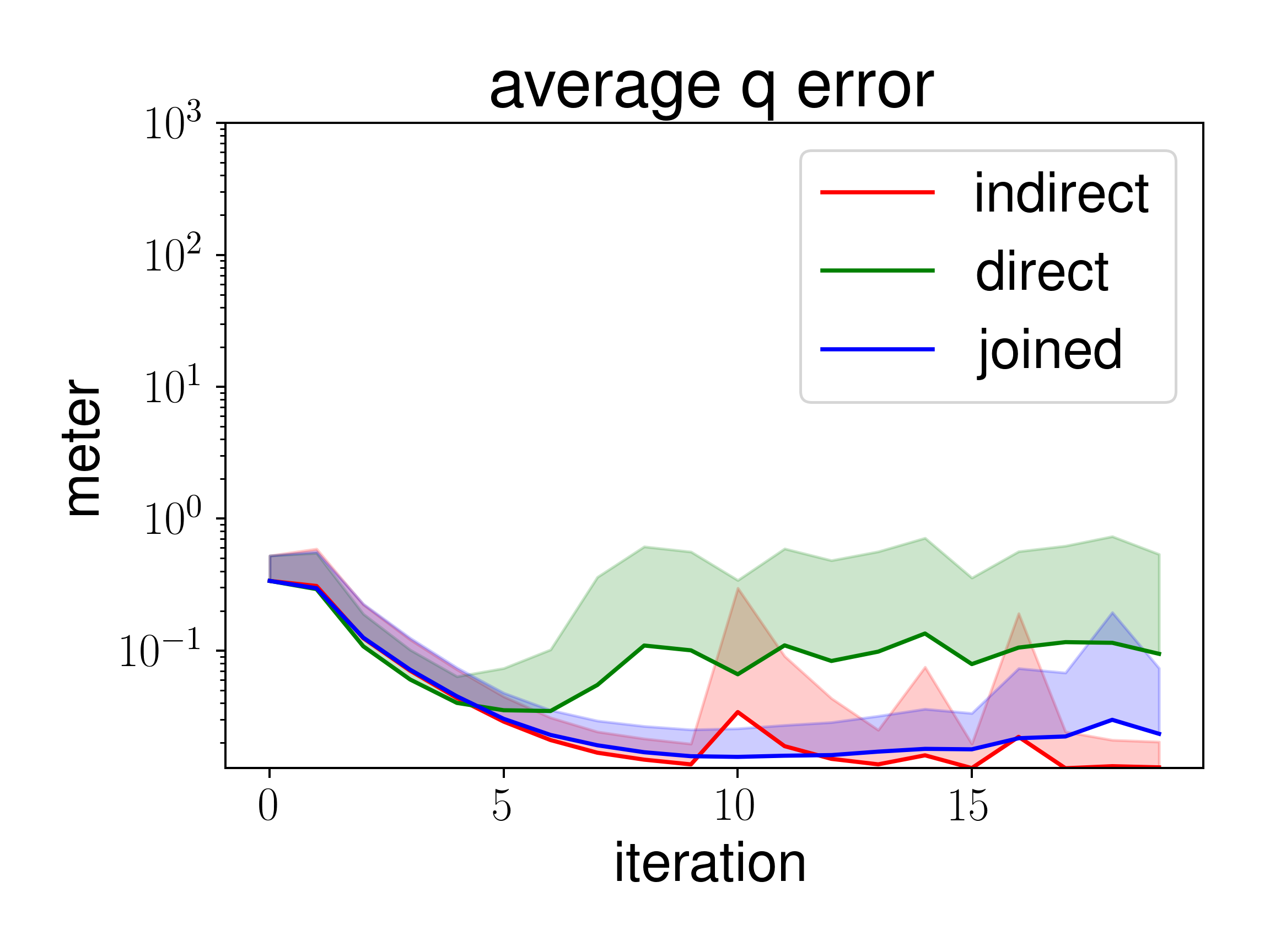}%
    \end{subfigure}
    \begin{subfigure}{.24\textwidth}
      \centering
      \includegraphics[width=0.99\linewidth]{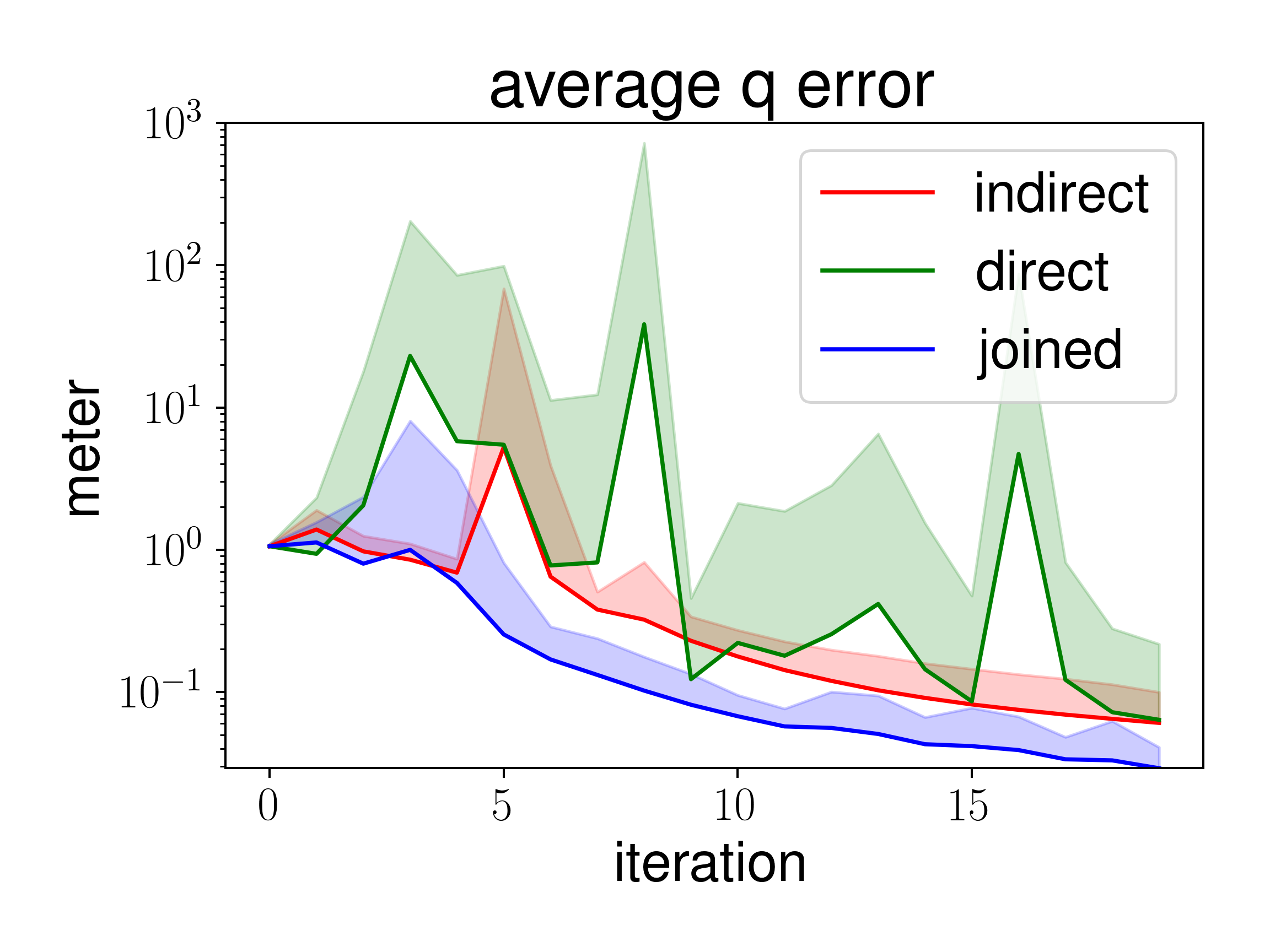}
    \end{subfigure}
    \begin{subfigure}{.24\textwidth}
      \centering
      \includegraphics[width=0.99\linewidth]{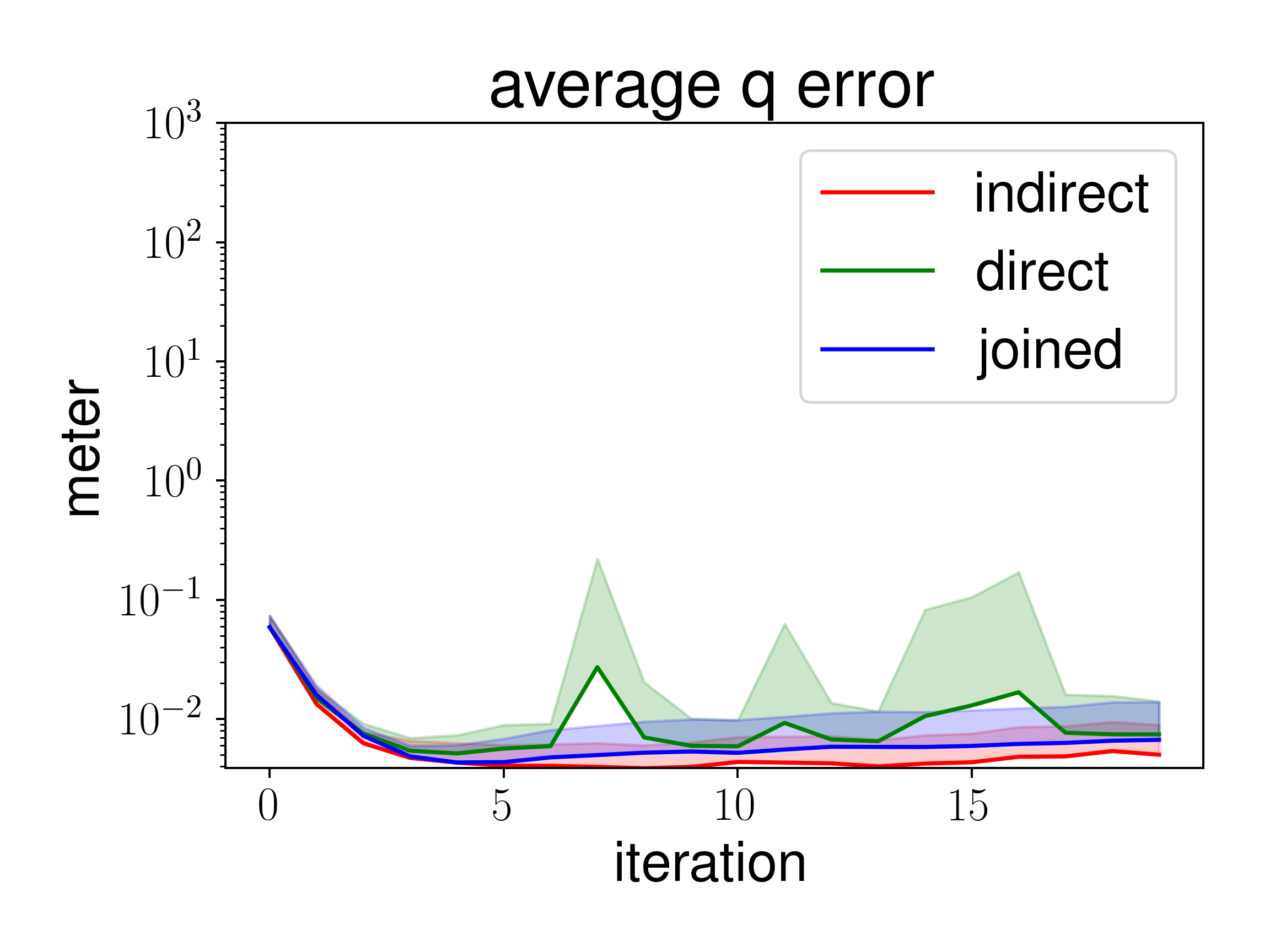}
    \end{subfigure}
    \begin{subfigure}{.24\textwidth}
      \centering
      \includegraphics[width=0.99\linewidth]{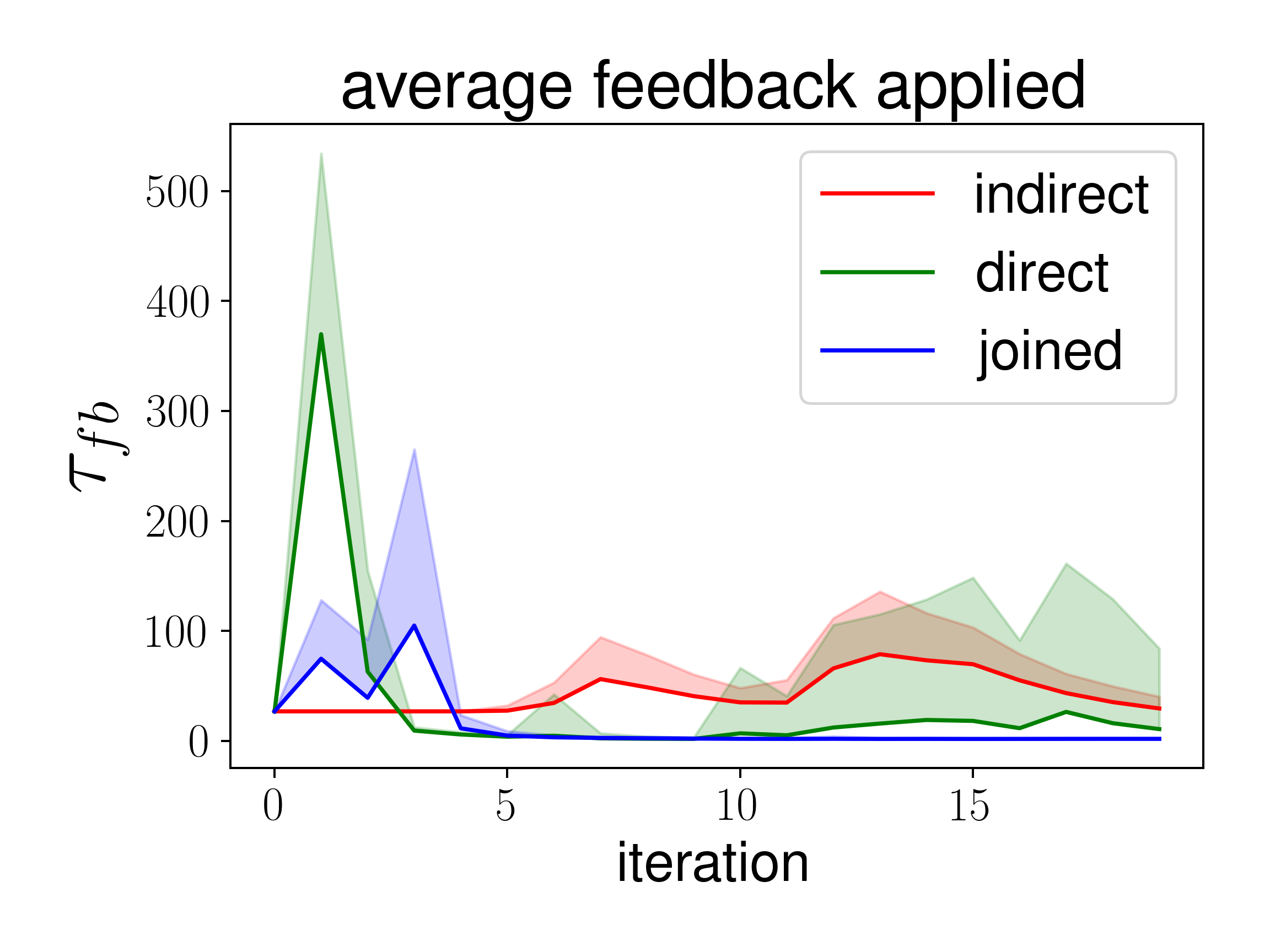}
      \caption{PID: low-gain}
    \end{subfigure}
    \begin{subfigure}{.24\textwidth}
      \centering
      \includegraphics[width=0.99\linewidth]{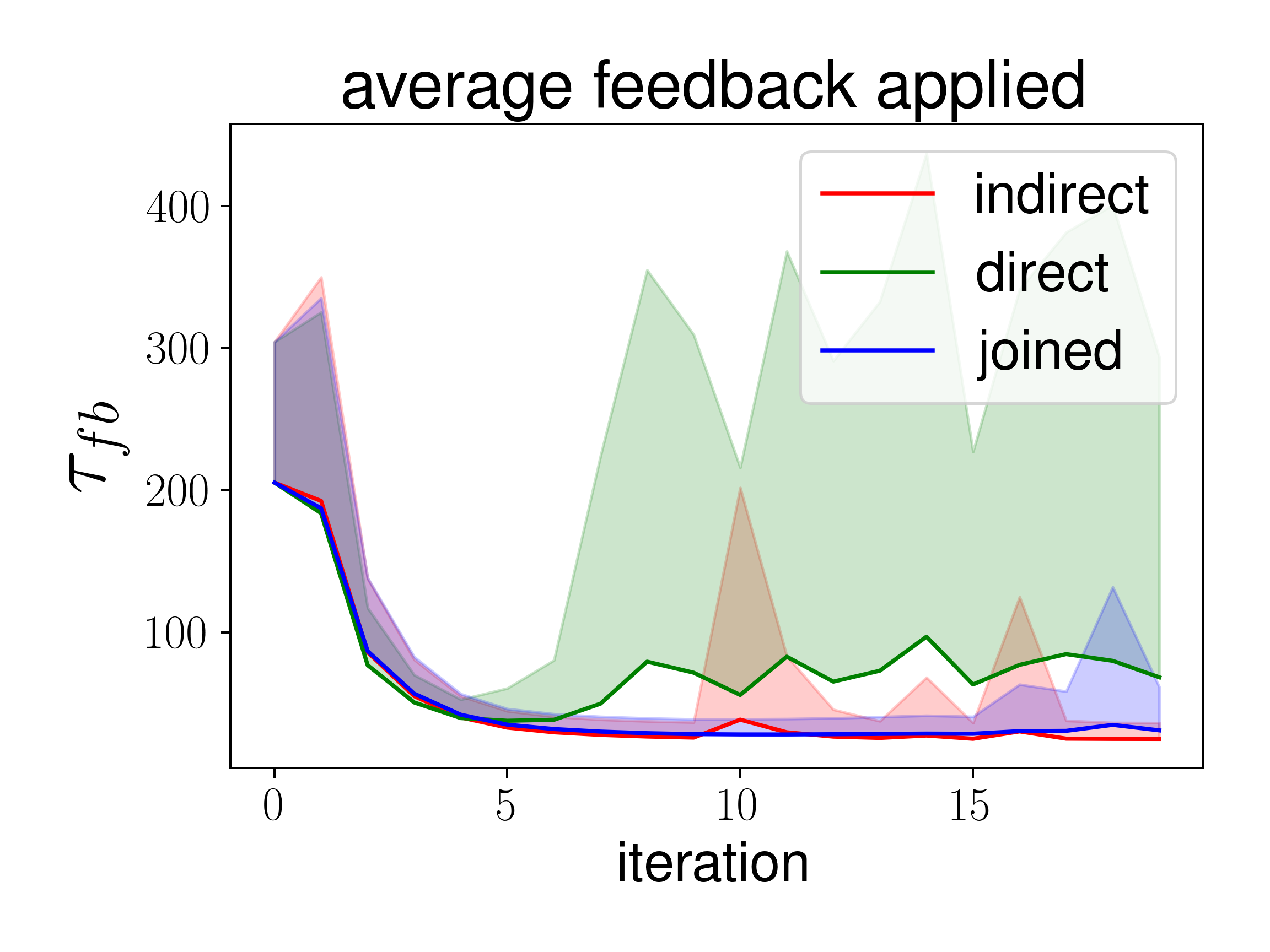}
      \caption{PID: high-gain}
    \end{subfigure}
    \begin{subfigure}{.24\textwidth}
      \centering
      \includegraphics[width=0.99\linewidth]{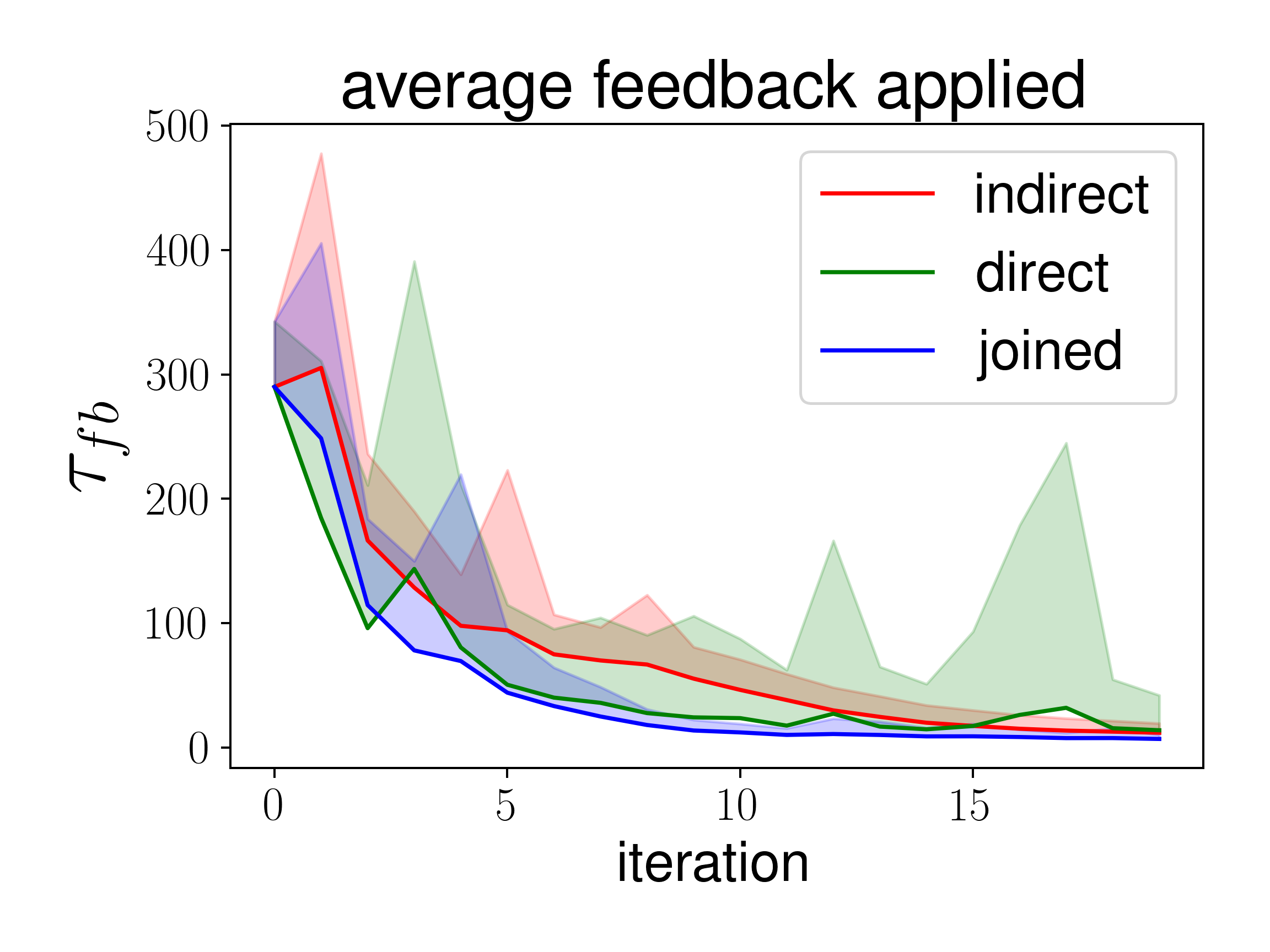}
      \caption{doomed: low-gain}
    \end{subfigure}
    \begin{subfigure}{.24\textwidth}
      \centering
      \includegraphics[width=0.99\linewidth]{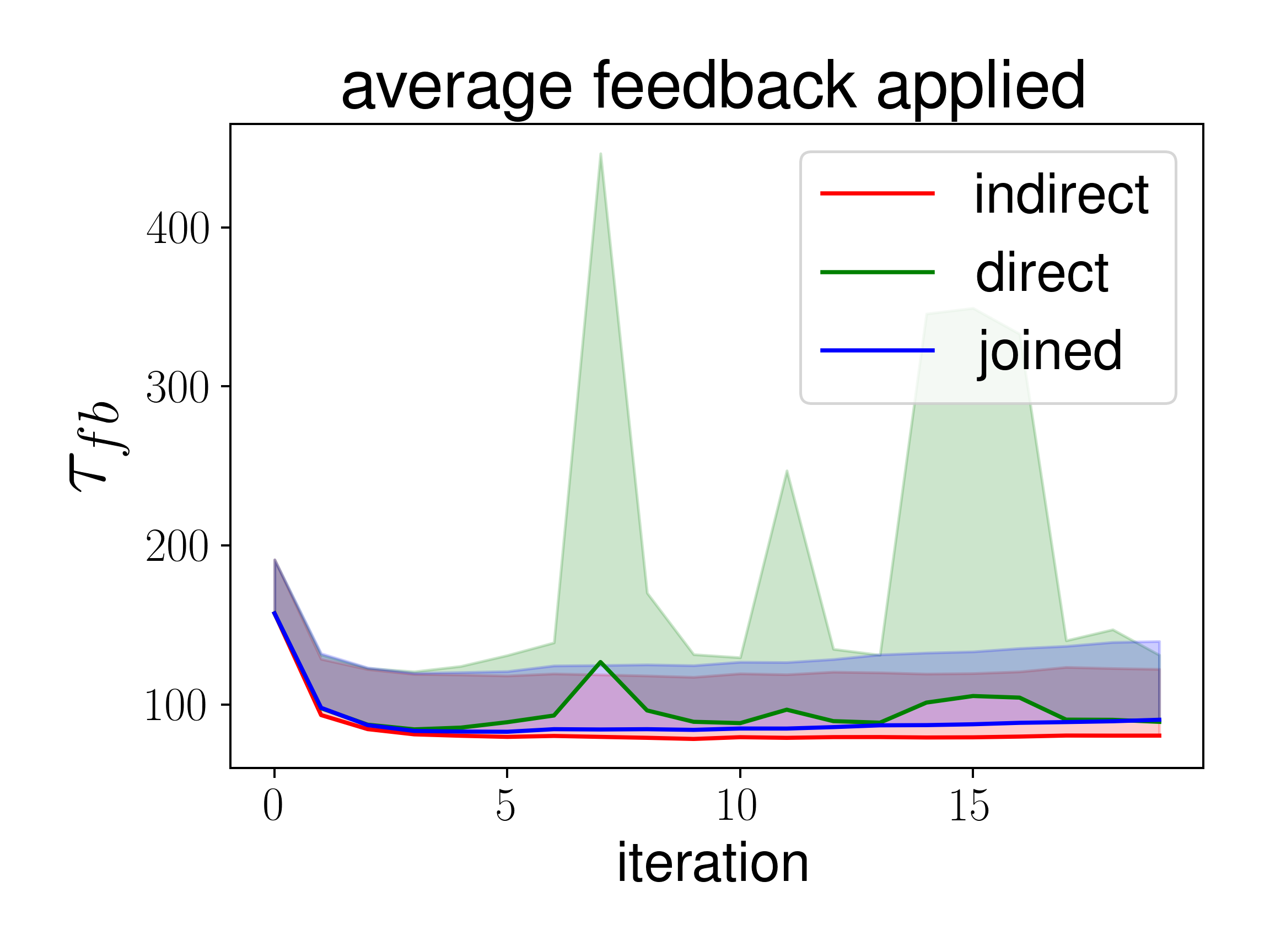}
      \caption{doomed: high-gain}
    \end{subfigure}
    \caption{\small Average results for low-gain and high-gain
      feedback control, and averaged across all system and parameter
      settings. (top row) shows the position tracking error
      convergence as a function of the number of learning iterations.
      (bottom row) shows the average feedback term applied. We plot
      the mean and the mean plus one standard deviation of the
      results. \emph{low-gain:} In this setting, even when averaging
      across all system settings and parameters, we observe similar
      position tracking behavior for the data source variants as in
      our hand-picked illustrations. Indirect data alone is not able
      to capture the error model, which also explains why higher
      feedback torques are required in this setting. Using direct
      data alone, results have a higher variance. On average the mean
      tracking behavior degrades again after a few learning iterations
      (PID) or is somewhat erratic (DOOMED). Using the joint data set
      results in the most consistent tracking error convergence for
      both feedback controllers. \emph{high-gain} When using high
      feedback gains, the joint data set method does not gain as much
      in convergence performance (over using indirect data alone).
      However using both data sources also does not degrade
      performance.}
    \vspace{-0.5cm}
	\label{fig:simulation_results_avg}
\end{figure*}

To provide a more extensive evaluation we now present results obtained
when averaging across all system and parameter settings, see
Fig.~\ref{fig:simulation_results_avg}. These results show that, on
average, using both data sources results in more consistent and faster
convergence of the position tracking error, when compared to using
the traditional approach of using indirect data alone.
Specifically, in the case of low-gain PID control, the error model trained with
the indirect data alone does not improve the position tracking error,
whereas the joint learning process does.

In the high-gain setting, the improvements are less pronounced.
However, we want to stress that one important goal of this work is to
learn an accurate task specific inverse dynamics model, while being as
compliant as possible.
Thus, the high gain setting shows that our proposed approach does not
deteriorate in case of a less compliant system configuration, but
there is not much to gain in using the additional data source.
Intuitively, this makes sense, since in a high gain setting, the
feedback control term is expected to provide good tracking performance
in the very first task execution already. Thus the indirect data
collected during that first run already provides very good data to
learn a model for that particular task.
We want to emphasize that there exist parameterizations and system
settings that can lead to better performance by a single data source.
However, on average, the direct data source seems to be most sensitive
to the system/parameter combination, and the indirect data source
requires higher feedback gains to be useful. The joint data
source, however, can achieve superior and more consistent model
learning performance, in low-gain settings.
\subsection{Real robot experiments}
\begin{figure}
    \centering
	\includegraphics[width=0.5\linewidth]{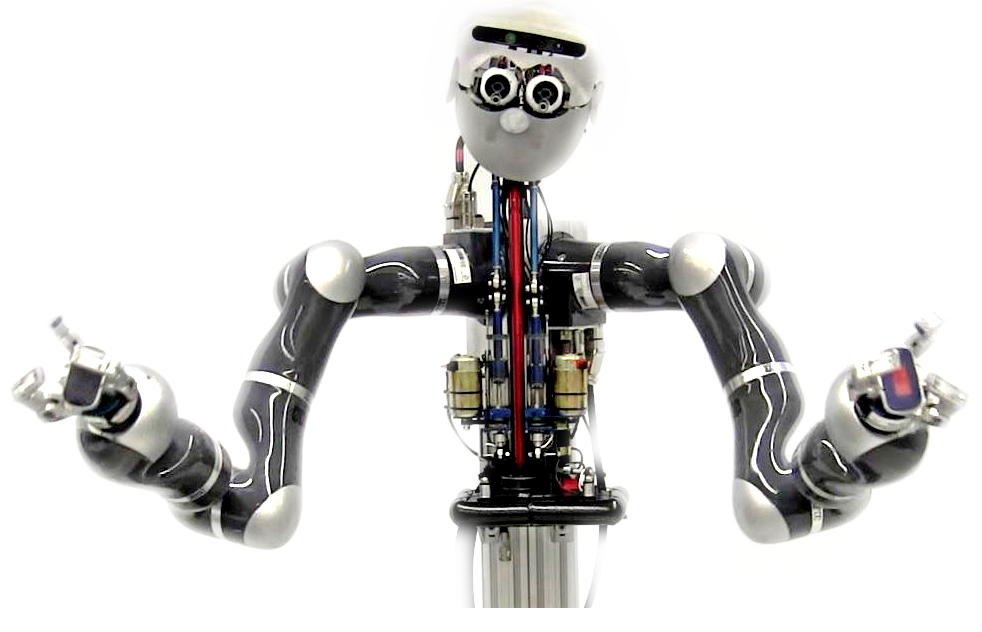}\\
	\includegraphics[width=0.65\linewidth]{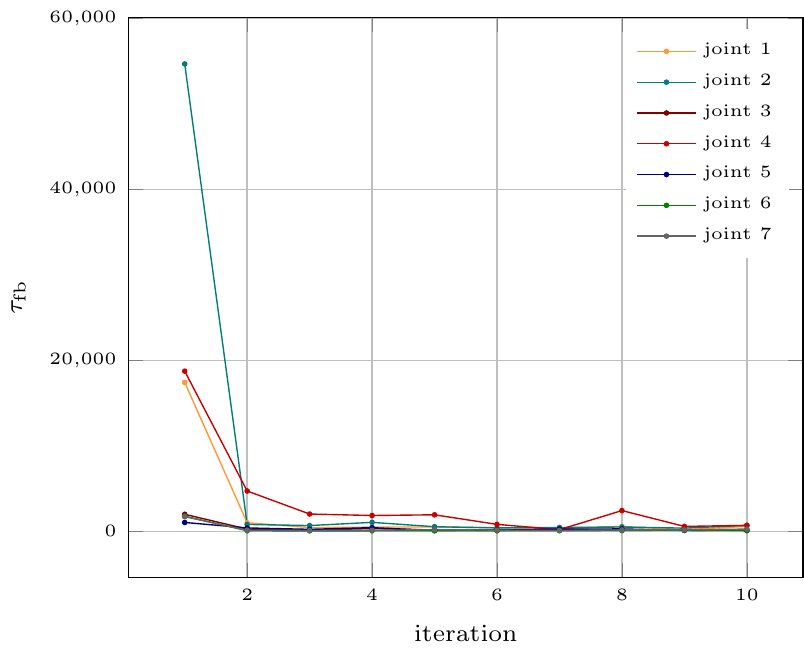}
	\caption{\small (top) Our robot platform is shown in the top
    figure. (bottom) Mean squared feedback terms ($\taufb$) as a
    function of learning iterations. Run 1 corresponds to
    Fig.~\ref{fig:apollo_result_error_offsets} (top) and run 10 to
    Fig.~\ref{fig:apollo_result_error_offsets} (bottom).
    }
	\label{fig:apollo_result_mse}
\end{figure}

We have evaluated our method with two real world robot experiments, a
quantitative and a qualitative one. Both experiments were
performed on the platform shown in Fig.~\ref{fig:apollo_result_mse} (top).
Our platform consists of two KUKA lighweight arms, each of which has $7$
degrees of freedom.
All experiments were performed on one arm resulting in a
$21$-dimensional input for the error model learning problem.
The control system operates on a hard real-time loop of 1 kHz, thus,
all predictions are performed in less than 1 ms. All our experiments
presented here, use DOOMED as feedback controller.
We optimize one neural network per joint with the following structure: 4 fully
connected layers (200, 100, 50, 1) with non-linearities (prelu
\cite{he2015delving}) after every layer except the last. Furthermore,
we bound the predicted torques to $\pm20 Nm$.
For both real robot experiments we analyzed the task-specific inverse
dynamics learning based on the joint data set, since this has shown 
to provide the most consistent performance.

\begin{figure}
    \centering
	\includegraphics[width=0.65\linewidth]{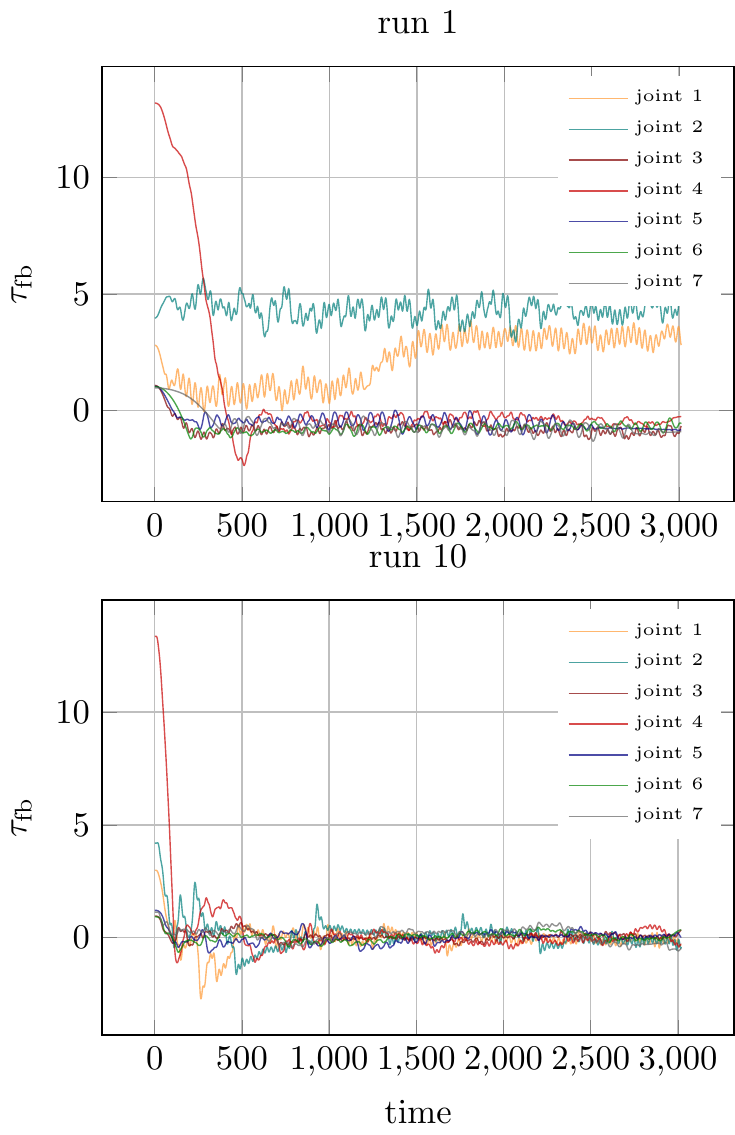}
	\caption{\small (top) The initial feedback terms ($\taufb$), per
    joint, without a learned error model. (bottom) $\taufb$ after 10
    learning iterations. }
	\label{fig:apollo_result_error_offsets}
\end{figure}

For our quantitative experiments we used a pre-planned sequence of LQR
policies to generate desired accelerations.
We execute the task 10 times, always starting in the same position, up
to the precision of the arm. After each run, the collected data was
used to re-optimize the neural networks.
Fig.~\ref{fig:apollo_result_mse} shows how the sum of squared feedback
terms $(\taufb)$, averaged across all joints, changes with each run.
Notice, the first run uses no error model, thus reflecting how much
the feedback term has to compensate for the modeling errors.
Within one iteration we are already able to capture most of the error
with the learned error model.
Hence, our learned model in combination with the rigid body dynamics
model is now a reliable inverse dynamics model for this task.
We want to stress that after every trial the newly obtained data is
used to further refine our task specific error model, and despite the data
correlation between trials our updated model does not degrade.
In Fig.~\ref{fig:apollo_result_error_offsets} we show the feedback
term trajectory for the first and last task execution, per joint.
Again it can be seen how our learned model compensates for the
errors such that the feedback term only has to adjust for
system noise.

In our qualitative example we show the data efficiency of our approach
for a real world manipulation task.
The robot has to pick up a heavy drill from a table and place it on a
different location on the same table.
Since we are interested in collaborative setting, we chose the
learning rate of DOOMED as low as possible such that the robot itself
is as compliant as possible while still being able to at least lift
the drill.
As shown in the video at \cite{sab_task}, the usage of the joint
data set, enables our system to significantly improve the performance
of the pick and place task after a single iteration.
	
%
\section{Discussion}
\label{sec:discussion}
%
In this work, we have proposed to use two different data sources to
learn inverse dynamics models. We have evaluated the usage of both
data sources, indirect and direct, both in simulation and on a real
system. Our evaluations demonstrate that combining the indirect and
direct data leads to more consistent and often faster learning
convergence, compared to using the traditional indirect data source
only. Furthermore, this superior performance of the combined data set
is especially noticeable in the low-gain feedback control setting,
such that we can effectively learn error models while being more
compliant from the beginning.

Nevertheless some restrictions to our approach exist: The proposed
method is based on the assumption that we do not visit the same part
of the state space $(\q, \qd)$, with differing amounts of payload.
Since the input of the error model does not contain any information
about the payload change of the system other then the generated
feedback $\taufb$, such a task would lead to data ambiguities.
For example, picking up a drill from a position on a table, placing it
somewhere else and repeating the pickup without the drill could not be
expressed by a single task-specific inverse dynamics model right now.
A potential extension to alleviate this problem would be to provide
additional inputs, either sensing, or more abstract provided
information to the learning system.

For this work we do not analyze how the system performs when it is
strongly perturbed.
This could result in undesirable predictions since the error model has
not been trained on any data for that input space.
This is however a general problem for task specific models. 
We want to emphasize that we have bounded the output of the error
model to a reasonable torque limit for our system, but this limit has
not been exceeded during our experiments.
Furthermore, we believe that this problem can be addressed in future
work as discussed in the following section.
\section{Conclusions and future work}
\label{sec:conclusion}
%
It is important to note, that in individual experiments, error models
trained on the direct, indirect and joint data have all shown
superior performance for certain system and learning
parameterizations.
Overall training on the joint data set results in more consistent and
faster convergence and lower exerted torques.
However, an interesting direction for future work is to exploit the
structure in the different data sources in order to identify which
source is more reliable.
Ideally this will allow to train an error model for which the
performance is always at least as good as the better one of the two
data sources.

We want to further investigate the scheduling of the gains of the
feedback term based on the performance of the error model.
This should enable our system to increase the level of compliance even
more over time. Being able to lower the gains also enables better
detection of structured perturbations, e.g. when a user is pushing the
robot arm.
The main reason for that is that our learned error model can capture
the modeling errors of the rigid body dynamics and the feedback term 
only has to correct for sensor noise.
Therefore, it is possible to detect otherwise difficult problems
such as collision with objects.

To the best of our knowledge the usage of the two data sources is a
novel approach to inverse dynamics learning.
We have empirically shown, that even with a low-gain feedback
controller, this can lead to consistent and fast error model learning.
The effectiveness of this approach has further been evaluated on a real system.

\section{Acknowledgments}
This research was supported in part by National Science Foundation
grants IIS-1205249, IIS-1017134, EECS-0926052, the Office of Naval
Research, the Okawa Foundation, and the Max-Planck-Society. Any
opinions, findings, and conclusions or recommendations expressed in
this material are those of the author(s) and do not necessarily
reflect the views of the funding organizations.

\bibliographystyle{IEEEtran}
{\scriptsize
\bibliography{IEEEabrv,abrv,references}
}

\end{document}